\documentclass[10pt,twocolumn,letterpaper]{article}

\usepackage[pagenumbers]{iccv} 

\definecolor{iccvblue}{rgb}{0.21,0.49,0.74}
\usepackage[pagebackref,breaklinks,colorlinks,allcolors=iccvblue]{hyperref}

\usepackage{algorithm} 
\usepackage{algorithmic}

\usepackage{multirow} 
\usepackage{colortbl}  
\definecolor{Gray}{gray}{0.94}
\newcommand{\shline}{\specialrule{.05em}{.05em}{.05em}}

\def\paperID{7906} 
\def\confName{ICCV}
\def\confYear{2025}

\title{Beyond Label Semantics: \\ Language-Guided Action Anatomy for Few-shot Action Recognition}

\author{
Zefeng Qian\textsuperscript{1} \quad 
Xincheng Yao\textsuperscript{1} \quad 
Yifei Huang\textsuperscript{3,4} \quad 
Chongyang Zhang\textsuperscript{1,2} \thanks{Corresponding author.} \quad 
Jiangyong Ying\textsuperscript{5} \quad 
Hong Sun\textsuperscript{5}  \\
\textsuperscript{1}School of Information Science and Electronic Engineering, Shanghai Jiao Tong University \\
\textsuperscript{2}MoE Key Lab of Artificial Intelligence, AI Institute, Shanghai Jiao Tong University \\
\textsuperscript{3}The University of Tokyo \quad
\textsuperscript{4}Shanghai AI Laboratory \quad
\textsuperscript{5}E-surfing Vision Technology Co., Ltd \\
{\tt\small \{zefeng\_qian, i-Dover, sunny\_zhang\}@sjtu.edu.cn, hyf@iis.u-tokyo.ac.jp} \\
{\tt\small \{yingjiangyong, sunhong\}@chinatelecom.cn} \\
}

\begin{document}
\maketitle
\begin{abstract}
Few-shot action recognition (FSAR) aims to classify human actions in videos with only a small number of labeled samples per category. 
The scarcity of training data has driven recent efforts to incorporate additional modalities, particularly text. 
However, the subtle variations in human posture, motion dynamics, and the object interactions that occur during different phases, are critical inherent knowledge of actions that cannot be fully exploited by action labels alone.
In this work, we propose \textbf{L}anguage-\textbf{G}uided Action \textbf{A}natomy (LGA), a novel framework that goes beyond label semantics by leveraging Large Language Models (LLMs) to dissect the essential representational characteristics hidden beneath action labels.
Guided by the prior knowledge encoded in LLM, LGA effectively captures rich spatiotemporal cues in few-shot scenarios.
Specifically, for text, we prompt an off-the-shelf LLM to anatomize labels into sequences of atomic action descriptions, focusing on the three core elements of action (subject, motion, object).
For videos, a Visual Anatomy Module segments actions into atomic video phases to capture the sequential structure of actions.
A fine-grained fusion strategy then integrates textual and visual features at the atomic level, resulting in more generalizable prototypes. 
Finally, we introduce a Multimodal Matching mechanism, comprising both video-video and video-text matching, to ensure robust few-shot classification. 
Experimental results demonstrate that LGA achieves state-of-the-art performance across multiple FSAR benchmarks.
\end{abstract}
\vspace{-2mm}

\section{Introduction}
\label{sec:intro}
Few-shot action recognition seeks to classify human actions in videos using only a limited number of labeled samples, addressing the challenge of data scarcity in real-world applications. Building on advances in few-shot image classification~\cite{One-shot2006, hariharan2017low}, many existing approaches adopt a metric-based meta-learning paradigm, where videos are embedded into a shared feature space, and alignment metrics are leveraged to infer the labels of the queried video~\cite{2020-OTAM}. The core challenge lies in learning effective representations that can transfer to novel, unseen action categories, while overcoming the inherent complexity of video data and the limited supervision available~\cite{HCL,STRM,2022-HyRSM,IST}. 


\begin{figure}[t]
  \centering
  \includegraphics[width=\linewidth]{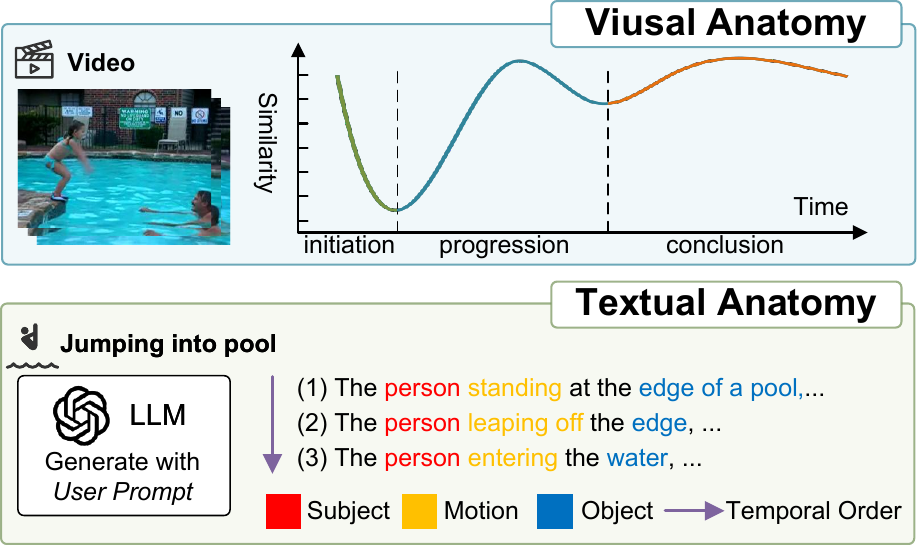}
   \caption{Illustration of our motivation. By leveraging the powerful knowledge understanding capability of LLMs, we anatomize one action label into a three-stage atomic action description. Meanwhile, the video is divided into corresponding three phases.}
   \label{fig:motivation}
   \vspace{-4mm}
\end{figure}

Recently, researchers have explored the integration of multimodal information to enhance the learned representations~\cite{AmeFu-Net,2023-molo,liu2023Lite-MKD,wang2023clip,xia2023few,shi2024commonsense}.
Among these, text has emerged as a particularly effective modality~\cite{wang2023clip, MORN, shi2024commonsense}. 
For example, CLIP-FSAR \cite{wang2023clip} regards action labels as text and incorporates CLIP~\cite{CLIP} to boost the FSAR performance. 
Shi \textit{et al.}~\citep{shi2024commonsense} constructs an external action knowledge base and aligns video frames with textual sub-action proposals.
%
However, human actions exhibit intricate variations and rich spatiotemporal details \cite{zhou2022thinking, ji2020actiongenome, li2020pastanet, sun2020discriminative}.
Subtle differences in posture, motion dynamics, and object interactions across different action phases serve as crucial knowledge for action recognition, yet these cues cannot be fully exploited by relying on simple semantic cues such as action labels.

To address this limitation, we first take a deeper look beyond the label semantics. 
Beyond the textual label, key factors of actions include the subject, motion dynamics, and object interactions~\cite{ji2020actiongenome,rai2021homeactiongenome}.Meanwhile, temporally, the way an action initiates, progresses, and concludes plays a crucial role in determining its category~\cite{zhao2017temporal, lin2019bmn,Gao_2022_FTCL}. Thus, achieving robust few-shot action recognition requires fine-grained alignment between query and support videos, ensuring that every critical aspect of the action is considered.

With this insight, in this work, we propose \textbf{L}anguage-\textbf{G}uided Action \textbf{A}natomy (LGA): a novel framework that anatomizes both textual and visual modalities to fully exploit their inherent knowledge for fine-grained query-support matching. 
LGA first uses a Large Language Model (LLM) to decompose action labels into a sequence of semantic sub-steps, as illustrated in \cref{fig:motivation}. 
We choose LLM due to its strong general world knowledge and instruction-following ability~\citep{mann2020language, raffel2020exploring}, enabling us to focus on the core elements of actions, \textit{i.e.}, subject, motion, and object. 
Simultaneously, a Visual Anatomy Module segments the video into distinct temporal phases, capturing the initiation, progression, and conclusion of the actions. 
A Fine-grained Multimodal Fusion Module then integrates the textual and visual features at the anatomized level, constructing modality-aligned action prototypes.
Finally, to ensure robust few-shot matching, we propose a Multimodal Matching Module. 
This module performs matching from two complementary perspectives: video-video matching compares query and support video features, while video-text matching aligns query video features with textual representations. 
By fully decomposing and aligning the anatomized knowledge of actions, LGA enables fine-grained multimodal understanding and significantly improves few-shot action recognition performance.

Experimental results on multiple benchmarks demonstrate that our method outperforms other FSAR methods. Also, we conduct extensive ablation studies and provide qualitative visualizations, both of which substantiate the effectiveness of our method. The contributions can be summarized as follows:
\begin{itemize}
\itemsep=0pt
\item We propose LGA, a novel method for FSAR that anatomizes textual and video modalities to fully exploit the inherent knowledge in actions.
\item We propose novel multimodal fusion and matching modules that effectively integrate atomic-level textual and visual features, enabling robust FSAR through both video-video and video-text matching.
\item Our method achieves state-of-the-art performance across five widely used benchmarks.
\end{itemize}
\section{Related work}

\subsection{Few-shot Action Recognition}
Few-shot learning aims to develop models that can generalize to novel categories with only a limited number of labeled samples. 
Few-shot action recognition deals with videos that contain complex spatio-temporal information, which requires not only a deep understanding of contextual semantics, but also a comprehensive modeling of spatio-temporal relationships.
Mainstream FSAR methods follow a ``metric-based meta-learning" paradigm, where a common embedding space is first learned through episodic training, and alignment metrics are designed to calculate the similarities between the query video and support videos.
For example, OTAM \cite{2020-OTAM} uses dynamic time warping to better align the support and query videos temporally. TARN \cite{TARN} and MPRE \cite{MPRE} both introduce temporal attention mechanisms to help the network generalize better to unseen categories. CPMT \cite{huang2022compound,huang2024matching} constructs multiple groups of prototypes to align each query against sequences in the support set and then fuses their matching scores for classification. 
HCL \cite{HCL} introduces a hierarchical matching model to support comprehensive similarity measures at global, temporal, and spatial levels via a zoom-in matching module.

While these methods have explored various temporal modeling and alignment strategies, they are fundamentally constrained by their reliance on single-modality representations. 
This leads to overfitting on seen categories and limits generalization to novel action classes in FSAR.
In our approach, we introduce the fine-grained fusion of visual and semantic features and multimodal classification between support videos and query videos, thereby significantly enhancing the model’s performance in few-shot scenarios.

\begin{figure*}[t]
  \centering
  \includegraphics[width=\textwidth]{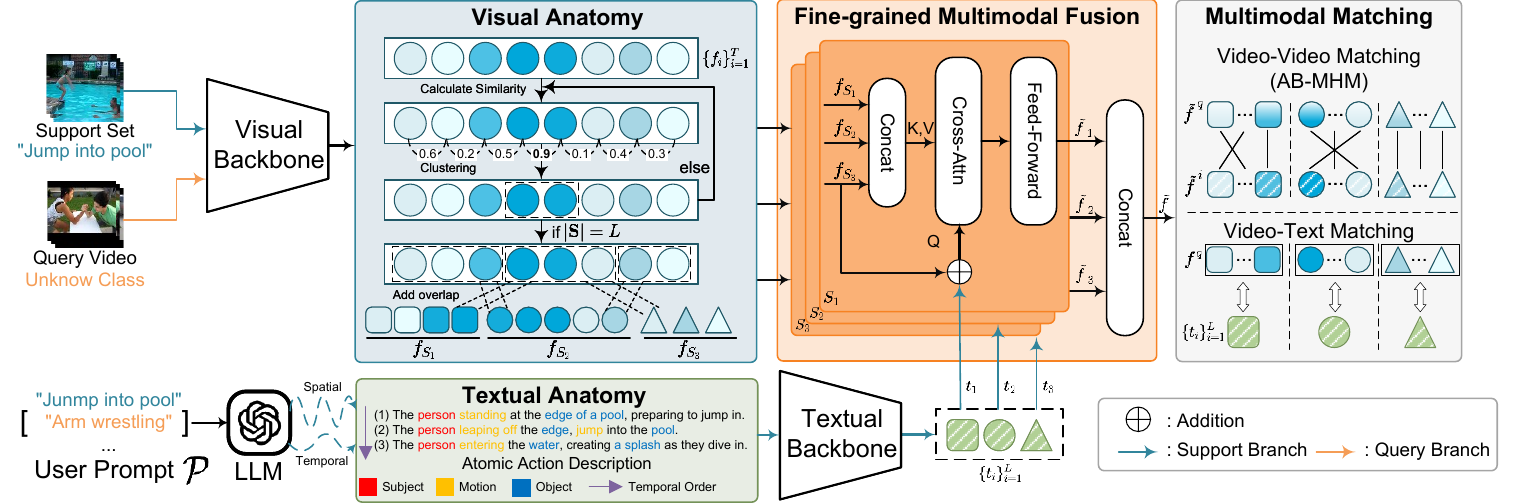}
   \caption{Illustration of our proposed method. First, the support and query videos are processed by the visual backbone to extract the visual features. Then, we transfer the action labels into atomic action descriptions with pre-trained large language model (LLM). Simultaneously, the Visual Anatomy Module segment frame sequence into atomic action phases. Then, we apply a Fine-grained Multimodal Fusion Module to integrate the visual features and semantic information at the atomic level, resulting in the final action prototype. Finally, we introduce a Multimodal Matching Module to achieve robust few-shot matching, which consists of both video-video matching and video-text matching.
    }
   \label{fig:framework}
   \vspace{-4mm}
\end{figure*}

\subsection{Multimdoal FSAR}
To alleviate the data scarcity issue, several methods attempt to enhance feature representation with the complementarity of different modalities in FSAR.
For example, AMeFu-Net~\cite{AmeFu-Net} proposes to use depth information for FSAR. 
MoLo~\cite{2023-molo} and AMFAR\cite{wanyan2023AMFAR} exploit the complementarity between visual features and motion cues.
Furthermore, Lite-MKD \cite{liu2023Lite-MKD} utilizes a teacher model to conduct multimodal learning to fuse optical flow, depth, and appearance features of human movements.


The integration of textual information shows great promise due to the high-quality representations provided by the pre-trained backbones such as CLIP~\cite{CLIP}.
CLIP-FSAR~\cite{wang2023clip} and MORN \cite{MORN} apply the pre-trained CLIP model to extract text and visual features, leveraging action labels as textual cues to enhance prototype learning.
Inspired by this, several methods \cite{xing2023maclip, cao2024task-adpter, guo2025clipcpmc, wu2025empnet, li2025TSAM, xing2025mafsar} adopt adapter-based approaches to balance computational efficiency and accuracy in FSAR.
Furthermore, Xia \textit{et al.} \cite{xia2023few} and Shi \textit{et al.} \cite{shi2024commonsense} construct external action knowledge bases from large-scale video captions and text corpora to enrich action representations.
SAFSAR \cite{Tang_2024_SAFSAR} extends elaborated descriptions from class names to better integrate textual and visual features in FSAR.
%

While these methods exploit multimodal cues, they fail to fully decompose and leverage the inherent structure of actions, leading to suboptimal prototype representations for novel action categories. 
In contrast, our proposed LGA framework addresses this gap with two key innovations:
(1) Previous approaches rely solely on action labels or externally collected text proposals, failing to fully exploit the inherent knowledge in the text and video of the actions. In contrast, we decompose action labels into a sequence of atomic action steps that explicitly encode the three core elements of actions (subject, motion, and object). Correspondingly, we anatomize the video into temporal segments of initialization, progression, and conclusion, yielding finer-grained textual and visual representations.
(2) Furthermore, our method aligns and fuses textual and visual features at the atomic level to generate robust prototypes, while previous CLIP-based FSAR methods neglect fine-grained alignment and fusion between different modalities.


\section{Method}
\subsection{Overview}
\textbf{Problem Definition.} The goal of few-shot action recognition (FSAR) is to classify an unlabeled query video into one of the $\mathnormal{N}$ action classes in the ``support set'' comprising $\mathnormal{K}$ labeled samples for each class that is unseen during training ($\mathnormal{N}$-way $\mathnormal{K}$-shot task). Following the typical ``metric-based meta-learning '' paradigm~\cite{snell2017prototypical}, we train a deep network on the training set $\mathcal{D}_{train}$ and then verify the model on the test set $\mathcal{D}_{test}$. All categories in the test set are not included in the training set.
As in previous approaches \cite{2023-molo,wang2023clip},
in the training phase, we adopt the episode-based meta-learning strategy to optimize the network. 
In each episode, the network is trained to classify query samples into correct categories, and optimized to assume that query samples are in the same category as the support samples closest to it. In the test phase, a large number of episodes are randomly selected from the test set, and the average accuracy is utilized to evaluate the few-shot performance of the trained model.

\noindent\textbf{Overall Architecture.} An overview of our framework is provided in \cref{fig:framework}, which mainly consists of three components: Action Anatomy, Fine-grained Multimodal Fusion, and Multimodal Matching. 
Following previous methods \cite{perrett2021trx, HCL,2023-molo}, we first sparsely random sample $T$ frame from the videos, and extract the image features with the feature extractor. 
We then anatomize actions in both textual and video modalities.
Specifically, for text, we leverage pre-trained LLM to transform each label into sequences of atomic action descriptions, focusing on the three core elements of action: subject, action, and object.
For videos, we decompose the visual features of the whole action into three sets of atomic action sequences, representing the initiation, progression, and conclusion phases.
Then, to enhance the generalization ability of the action prototype, we use a Fine-grained Multimodal Fusion Module to integrate the anatomized visual and textual features, obtaining the action prototypes.
Finally, based on the action prototypes, we leverage a novel Multimodal Matching Module to achieve robust few-shot matching. The Multimodal Matching consists of video-video matching and video-text matching.
We introduce the details of each component below.

\subsection{Action Anatomy}
\label{sec:action anatomy}
The first key step in LGA is action anatomy, which is conducted on both the action labels and the videos. The anatomized text and videos are then fused and matched by subsequent modules for few-shot action recognition. 

\noindent\textbf{Textual Anatomy.}
To harness the rich prior knowledge embedded in LLM, we decompose each action label into an ordered sequence of atomic action descriptions that capture fine-grained spatiotemporal context. 
%
Specifically, we design a carefully crafted prompt that guides the LLM to decompose an action label into a sequence of semantically meaningful sub-actions, ensuring that the subject, motion, and object are explicitly represented at each stage. The generated descriptions follow the natural flow of action: from \textit{initiation} to \textit{progression} and finally to \textit{conclusion}, allowing for a more structured and expressive textual representation.

An example of the anatomy of the label ``Jump into pool" can be seen in ~\cref{fig:framework}. The LLM generates ``A person standing at the edge of a pool, preparing to jump in.", ``The person leaping off the edge, jump into the pool." and ``The person entering the water, creating a splash as he dives in." 
These fine-grained atomic-level action descriptions provide explicit spatiotemporal cues compared to the action label.
Finally, we feed the atomic action descriptions into the textual backbone to extract features $\{t_i\}_{i=1}^L$, where $L$ is the number of the generated atomic descriptions.


\noindent\textbf{Visual Anatomy:}
The Visual Anatomy Module transforms sequences of frame features into sets of atomic phases corresponding to the generated atomic action descriptions.
As shown in \cref{fig:framework}, given a video sequence $\{f_i\}_{i=1}^T$, we transform it into distinct groups of atomic action sequences $\{f_{S_1}, f_{S_2}, ... ,f_{S_L}\}$, where $L$ represents the number of atomic actions in the video, corresponding to the atomic action textual descriptions.

To accurately anatomize the visual sequence feature into atomic phases, we propose a CLUSTER-Segment strategy that identifies sub-actions of varying temporal durations. This strategy encourages the frames within the same sub-action to be similar, while ensuring that frames belonging to different sub-actions remain as distinct as possible.
Specifically, for an initial frame sequence $\{f_i\}_{i=1}^T$, we treat each frame as an individual cluster $\{S_i\}_{i=1}^T$. We then compute the cosine similarity between adjacent clusters, and merge the pair of adjacent clusters exhibiting the highest similarity to obtain the updated cluster $\{S_i\}_{i=1}^{T-1}$. This merging process is repeated until only $L$ clusters remain. Additionally, to enhance the robustness of the CLUSTER-Segment algorithm, we introduce duplicated frames within adjacent clusters. More details are described in \cref{algo_cluster}.

\begin{algorithm}
\caption{Algorithm for CLUSTER-Segment in Visual Anatomy Module}
\label{algo_cluster}
\begin{algorithmic}
\REQUIRE frame feature $\{f_i\}_{i=1}^T, f_i\in \mathbb{R}^C $
\STATE initialize cluster $\mathbf{S} = \{S_i\}_{i=1}^T$,\\ target number of atomic actions of $L$
\WHILE{$\lvert\mathbf{S}\rvert > L$}
    \STATE $\mathbf{L}_{sim}$ = []
    \STATE $\overline{f}_{S} = \frac{1}{|S_i|} \sum f_i \quad \text{where } i \in S_i$
    \FOR{$i = 0$ \TO $\lvert\mathbf{S}\rvert$}
        \STATE $\mathrm{Sim}$ = \text{CosineSimilarity}$(\overline{f}_{S_i},\overline{f}_{S_{i+1}})$
        \STATE \# Append $\mathrm{Sim}$ to $\mathbf{L}_{sim}$
    \ENDFOR
    \STATE $\mathrm{j}$ =  $\operatorname{argmax}(\mathbf{L}_{sim})$
    \STATE \# Merge cluster $\mathbf{S_j}$ and $\mathbf{S_{j+1}}$
\ENDWHILE
\STATE \# Add overlap frame within adjacent clusters 
\RETURN Temporal-Segment Result $\mathbf{S} = \{S_i\}_{i=1}^L$
\end{algorithmic}
\end{algorithm}

\subsection{Fine-grained Multimodal Fusion}
\label{sec:Fine-grained Multimodal Fusion}
After anatomizing both visual and textual features, the next challenge is to effectively integrate these two modalities to construct a robust action representation.
To this end, we propose a Fine-grained Multimodal Fusion Module designed to align and fuse visual and semantic information at the anatomized action level, as illustrated in \cref{fig:framework}.

In Fine-grained Multimodal Fusion Module, given the anatomized visual feature $\left\{f_{S_i} \;\middle|\; S_i \in \{S_1, \dots, S_L\} \right\}$ and its corresponding textual feature $\{t_i\}_{i=1}^{L}$, our goal is to establish fine-grained correspondences between these representations. Specifically, we employ a multi-head Cross-Attention mechanism to facilitate interaction between modalities.
In our design, we formulate the query $\mathbf{Q}$ as the sum of the atomic visual feature $f_{S_i}$ and its aligned textual feature $t_i$.
Meanwhile, the key and value matrices $\mathbf{K}$ and $\mathbf{V}$ are constructed by concatenating all atomic visual features.
This design ensures that each atomic action feature not only learns localized semantic details but also maintains awareness of the global temporal structure of the action.
Subsequently, a feed-forward network processes the cross-attended information to generate the fused feature $\tilde{f}_{S_i}$ for the atomic action. 
Finally, we concatenate the atomic action features across all temporal phases to construct the final action prototype $\tilde{f}$.
The process can be formulated as:
\begin{equation}
\small
\begin{aligned}
    \mathbf{Q_i} &=  t_i+f_{S_i},  \mathbf{K} = \mathbf{V} = \operatorname{concat}(\{f_{S_i}\}_{i=1}^L), \\
    f_{S_i}&=\operatorname{Softmax}\left({\frac{\mathbf{Q_i}\mathbf{K}^{\top}}{\sqrt{d_{k}}}} \right)\mathbf{V}, \tilde{f}_{S_i} = \mathbf{FFN}(f_{S_i})+f_{S_i}, \\
     \quad \tilde{f} &=  \operatorname{concat}(\tilde{f}_{S_1}, \tilde{f}_{S_2}, \tilde{f}_{S_3}).
\end{aligned}
\end{equation}

We then show how we leverage the action prototype for robust few-shot matching.

\subsection{Multimodal Matching}
In few-shot scenarios, accurate classification depends on more than just measuring the distance between query and support videos based solely on visual feature matching. Previous methods often overlook the use of multimodal information, which limits their ability to handle the richness of both visual and textual cues~\cite{ITANet,perrett2021trx}.
Leveraging our anatomized representations and fused prototypes, we propose a Multimodal Matching Module that enhances classification accuracy by considering both video-video matching and video-text matching.

\noindent\textbf{Video-video matching:}
In video-video matching, we measure the distance between the features of the query video and the prototype features of each category within the support set.
In this context, it is crucial to account for the ordered temporal alignment between atomic actions. Existing methods, such as the bidirectional Mean Hausdorff Metric (Bi-MHM) \cite{2022-HyRSM}, typically operate on a global temporal scale and do not incorporate the inherent order of atomic actions across the video sequence. This approach fails to recognize that actions consist of temporally ordered sub-actions, and the distance between identical actions should be evaluated based on the alignment of these sub-actions at each time step.


To address this gap, we propose a novel metric \textbf{A}ligned \textbf{B}idirectional \textbf{M}ean \textbf{H}ausdorff \textbf{M}etric (AB-MHM), which explicitly incorporates ordered temporal alignment of atomic actions:
\begin{equation}
\begin{aligned}
    \tilde{\mathcal{D}}(\tilde{f}^q,\tilde{f}^i) = \frac{1}{T} \sum_{k=1} ^{L}[ &\sum_{\tilde{f}^q_m \in \tilde{f}_{S_{k}}^q}(\operatorname*{min}_{\tilde{f}_{n}\in\tilde{f}_{S_{k}}}\left|\left|{\tilde{f}^q_m-\tilde{f}^i_{n}}\right|\right|) \\
    + &\sum_{\tilde{f}_n \in \tilde{f}_{S_{k}}}(\operatorname*{min}_{\tilde{f}_{m}^{q}\in\tilde{f}_{S_{k}}^{q}}\left|\left|{\tilde{f}_n^i-\tilde{f}_{m}^{q}}\right|\right|)],\\
\end{aligned}
\end{equation}

\begin{equation}
    {p_{(y=i|q)}^{v-v}} = \operatorname{Softmax}(-\tilde{\mathcal{D}}(\tilde{f}^q,\tilde{f}^i)),
\end{equation}
where $T$ denotes the frame number. Finally, we take the negative distance for each class as the logit, and employ a softmax function to derive the probability $p^{v-v}$ that the query video belongs to category, thereby obtaining the classification results for video-video matching.

The key distinction of AB-MHM is that it aligns atomic actions in time, ensuring a more accurate measure of similarity. This approach not only improves the accuracy of video-video matching but also increases computational efficiency. AB-MHM is non-parametric and does not require complex alignment classifiers, making it more transferable and less computational costly. The benefits of incorporating ordered temporal alignment are demonstrated in \cref{sec:ablation}, where our method shows significant improvements over traditional matching approaches.


%
\begin{table*}[t]
\centering
\small
\caption{Comparison with state-of-the-art few-shot action recognition methods on HMDB51, Kinetics, UCF101, SSv2-Full, and SSv2-Small datasets with 5-way k-shot setting.
%
The best results are \textbf{bolded} in black and the \underline{underline} represents the second best result.
%
}

\setlength{
    \tabcolsep}{
    1.5mm}{
\begin{tabular}
{l|c|c|cc|cc|cc|cc|cc}
\hline
			
 \multirow{2}{*}{Method} \hspace{2mm} 
& \multicolumn{1}{c|}{\multirow{2}{*}{Reference}} 
& \multicolumn{1}{c|}{\multirow{2}{*}{Pre-training}} 
& \multicolumn{2}{c|}{{HMDB51}}  &\multicolumn{2}{c|}{{Kinetics}}  
& \multicolumn{2}{c|}{{UCF101}}  & \multicolumn{2}{c|}{{SSv2-Small}}
& \multicolumn{2}{c}{{SSv2-Full}}
\\
& \multicolumn{1}{c|}{} 
& \multicolumn{1}{c|}{} 
& \multicolumn{1}{l}{1-shot} & 5-shot
& \multicolumn{1}{l}{1-shot}  & 5-shot
& \multicolumn{1}{l}{1-shot} & 5-shot   
& \multicolumn{1}{l}{1-shot} & 5-shot  
& \multicolumn{1}{l}{1-shot} & 5-shot  \\
\shline
%
%
%
%
%
OTAM~\citep{2020-OTAM}    & CVPR'20  & INet-RN50
& 54.5  & 68.0   
& 73.0  & 85.8   
& 79.9  & 88.9   
& 36.4  & 48.0   
& 42.8  & 52.3   
\\

%
TRX~\citep{perrett2021trx}    & CVPR'21  & INet-RN50
& 53.1    & 75.6  
& 63.6  & 85.9   
& 78.2  & {96.1} 
& 36.0   &  {56.7}   
& 42.0  & 64.6   
\\
%
%
STRM~\citep{STRM}    & CVPR'22  & INet-RN50
& 52.3  &77.3     
& 62.9  & 86.7    
& 80.5 & {96.9}   
& 37.1 & 55.3     
& 43.1 & 68.1     
\\
HyRSM~\citep{2022-HyRSM}    &  CVPR'22  & INet-RN50
& 60.3  & 76.0     
& 73.7    & 86.1   
& 83.9  & {94.7}   
& 40.6  & 56.1     
& 54.3   &  69.0  
\\
%
%
%
CMPT~\citep{huang2022compound}    & ECCV'22  & INet-RN50
& 60.1  & 77.0    
& 73.3  & 86.4    
& 71.4  & 91.0    
& 38.9  & 61.6    
& 49.3  & 66.7    
\\ 
%
%
%
MoLo~\citep{2023-molo}    & CVPR'23   & INet-RN50
& {60.8}   & 77.4      
& {74.0}   & 85.6      
& {86.0}   & {95.5}    
& {42.7}   & {56.4}    
& {56.6}   & 70.6      
\\

GgHM \citep{GgHM} & ICCV'23 & INet-RN50
& 61.2  & 76.9   
& 74.9  & 87.4   
& 85.2  & 96.3   
& -     &  -     
& 54.5  & 69.2   
\\

Huang \textit{et al.}~\citep{huang2024matching}    & IJCV'24   & INet-RN50
& 61.6   & 77.5     
& 74.0   & 86.9     
& 74.9   & 92.5     
& 42.6   & 61.8     
& 52.3   & 67.1     
\\

TEAM ~\citep{lee2025TEAM}    & CVPR'25   & INet-RN50
& 62.8    & 78.4  
& 75.1    & 88.2  
& 87.2    & 96.2  
& -       & -
& -       &  -
\\

\shline
OTAM~\cite{2020-OTAM} & CVPR'20  & CLIP-ViT-B
& 72.5   & 83.9     
& 88.2   & 94.8     
& 95.8   & {98.8}     
& 43.3   & 57.5     
& 50.2   & 68.6     
\\  

CLIP-Freeze~\citep{CLIP} & ICML'21  & CLIP-ViT-B
& 58.2   & 77.0    
& 78.9   & 91.9    
& 89.7   & 95.7    
& 29.5   & 42.5    
& 30.0   & 42.4    
\\  

CapFSAR~\citep{CapFSAR} & ArXiv'23 & BLIP-ViT-B
&65.2  & 78.6   
&84.9  & 93.1   
&93.3  & 97.8   
&45.9  & 59.9   
&51.9  & 68.2   
\\  






MVP-shot~\citep{mvp-shot}& Arxiv'24 & CLIP-ViT-B
& 77.0   & \underline{88.1}
& 91.0  & 95.1 
& 96.8  & \underline{99.0}
& 55.4  &  62.0
& -     &   -  
\\


EMP-Net ~\citep{wu2025empnet}& ECCV'24 & CLIP-ViT-B
& 76.8    &  85.8   
& 89.1    & 93.5    
& 94.3    & 98.2    
& \underline{57.1}    &  \underline{65.7}    
& \underline{63.1} & \underline{73.0}  
\\  


CLIP-CPM$^2$C~\citep{guo2025clipcpmc}& Nc'25 & CLIP-ViT-B
& 75.9  & {88.0}   
& {91.0}  & \underline{95.5}   
& 95.0  & 98.6   
& 52.3  & 62.6   
& 60.1  & 72.8   
\\  

\hline
CLIP-FSAR~\cite{wang2023clip} & IJCV'23  & CLIP-ViT-B
& \underline{77.1}  & 87.7    
& \underline{94.8}  & 95.4    
& \underline{97.0}  & \textbf{99.1}    
& 54.6  & 61.8    
& 62.1  & 72.1    
\\  

\rowcolor{Gray}
Our~ & -  & CLIP-ViT-B
& \textbf{86.8}  & \textbf{89.3 }   
& \textbf{95.2}  & \textbf{96.2}  
& \textbf{98.2}  & \textbf{99.1}   
& \textbf{58.9}  & \textbf{69.3}  
& \textbf{63.8}    & \textbf{{74.4}   } 
\\

\hline
\end{tabular}
}
\label{tab:compare_SOTA_all}
\vspace{-2mm}
\end{table*}

\noindent\textbf{Video-text Matching:}
In video-text matching, we classify the query video by measuring the distance between its feature representation and the atomic-level textual features of each category.
Specifically, given a query feature ${\tilde{f}^q}$, we first apply average pooling to obtain the feature in each action phase $\overline{f}^q_{S_k}$.
Then, we calculate the similarity between the visual and semantic features at each action phase and classify the query video using a softmax function, which can be expressed as:
\begin{equation}
\begin{aligned}
    {p_{(y=i|q)}^{v-t}} &= \operatorname{Softmax}(\sum_{k=1}^L \langle \overline{f}^q_{S_k}, {f^{T}_k}\rangle))
\end{aligned}    
\end{equation}

Finally, we integrate the classification results from video-video matching and video-text matching using a weighted geometric mean as in \cite{wang2023clip}, which can be formulated as:
\begin{equation}
    {p}_{(y=i|q)} = \left(p_{(y=i|q)}^{v-v}\right)^\alpha \times {\left(p_{(y=i|q)}^{v-t}\right)}^{(1-\alpha)}
\end{equation}
where $\alpha \in [0,1]$ is an adjustable hyperparameter.

During the training phase, the entire framework is trained end-to-end, and we utilize the cross-entropy loss \cite{2020-OTAM, perrett2021trx} and contrastive loss \cite{2023-molo,2024-CCLN} to optimize the model.  
The contrastive loss enforces consistency in feature representations by contrasting positive and negative samples.
Notably, to ensure training stability, we adopt the results from video-video matching as the classification results during the training phase. 
During inference, we adopt the results of Multimodal Matching as the classification results for the query video.

\section{Experiments}

\subsection{Experimental setup}
\textbf{Datasets:}
We perform our experiments on five widely used few-shot action recognition datasets: SSv2-Full \cite{goyal2017ssv2}, SSv2-Small \cite{goyal2017ssv2}, Kinetics \cite{carreira2017kinetics}, UCF101 \cite{soomro2012ucf101}, and HMDB51 \cite{kuehne2011hmdb51}. 
For a fair comparison with previous methods, we follow the dataset split as described in \cite{2020-OTAM,2018-CMN}.

\noindent\textbf{Training and Inference:}
Following previous methods \cite{perrett2021trx,2022-HyRSM,2023-molo, wang2023clip}, we uniformly sample 8 frames (\textit{i.e.}, $T = 8$) to represent each video, and employ a ViT-B/16 \cite{2021vit} backbone with CLIP \cite{CLIP} initialization to serve as the visual and textual encoder.
In the training stage, our method is optimized end-to-end using the Adam optimizer \cite{kingma2014adam}.
In the inference stage, we centrally crop a $224 \times 224$ region from each frame. Following previous methods \cite{STRM,2022-HyRSM, 2023-molo}, we calculate the average classification accuracy of the 10,000 episodes from the test set to evaluate the few-shot performance on each benchmark.

\subsection{Comparison with state-of-the-art methods}
In this section, we use CLIP-FSAR~\cite{wang2023clip} as a baseline method and comprehensively compare our method with state-of-the-art approaches~\cite{HCL,2022-HyRSM,huang2022compound,huang2024matching,2023-molo,wang2023clip,wu2025empnet} using mainstream backbones (INet-RN50~\cite{2016-resnet} or ViT-B/16~\cite{2021vit}). For a fair comparison, we exclude task-specific methods~\cite{wang2023task,cao2024task-adpter} that leak task information, which compromise the transferability of the model.

As shown in \cref{tab:compare_SOTA_all}, we can make the following observations:
(a) Compared to the INet-RN50~\cite{2016-resnet} backbone, the methods with the CLIP-ViT-B~\cite{2021vit} backbone show superior performance. This indicates that the transformer architecture and pre-training contribute significantly to the performance.
(b) Compared to the baseline (CLIP-FSAR~\cite{wang2023clip}), our method achieves superior performance across all datasets, demonstrating its effectiveness in different scenarios.
(c) Compared to other methods, our approach also achieves competitive performance. Notably, the experiments show that our method delivers larger performance improvements on Kinetics and HMDB51. 
We attribute this to the fact that, compared to the SSv2 dataset, these datasets contain rich action-related elements that align well with our atomic action descriptions. 
For UCF101, the training data of the backbone already cover the action categories, leaving little room for further improvement.


\subsection{Ablation study}
\label{sec:ablation}

\begin{table}[t]
\centering
\small
\caption{Ablation study on HMDB51 datasets under 5-way 1-shot and 5-way 5-shot settings. The top two line represents the baseline CLIP-FSAR. ``Visual-An'' and ``Textual-An''
mean anatomize the video and action label into atomic units, respectively.}
\setlength{
    \tabcolsep}{
    1.0mm}{
\begin{tabular}
{cc|cc|cc}

\hline
			
\multicolumn{2}{c|}{Fusion} & 
\multicolumn{2}{c|}{Macthing}  &

\multicolumn{2}{c}{{HMDB51}}    \\
 \multicolumn{1}{l}{Visual-An} & \multicolumn{1}{l|}{Textual-An} &\multicolumn{1}{l}{video-video}  & \multicolumn{1}{l|}{video-text} & \multicolumn{1}{l}{1-shot} & \multicolumn{1}{l}{5-shot} \\ 
\shline

 & & \checkmark& & {75.8} & {87.7}  \\ 
& & \checkmark& \checkmark & {77.1} & {87.7}  \\ 
\hline 
\checkmark&   & \checkmark& & {79.9} & {86.0} \\ 
& \checkmark  & \checkmark& & {79.6} & {87.2} \\ 
\checkmark  & \checkmark & \checkmark& & {80.8} & {88.2} \\ 
\hline 
\checkmark&   & \checkmark& \checkmark & {83.1} & {86.2} \\ 
& \checkmark  & \checkmark& \checkmark & {85.2} & {87.8} \\ 
\rowcolor{Gray}
\checkmark& \checkmark &\checkmark & \checkmark & \textbf{{86.8}} & \textbf{{89.3}} \\ 

\hline

\end{tabular}
}
\label{tab:network_component_ablation}
\vspace{-2mm}
\end{table}
\noindent\textbf{Analysis of network components.}
In \cref{tab:network_component_ablation}, we perform ablation studies in 5-way 1/5-shot settings to investigate the effectiveness of different components in our methods. 
We can find that anatomizing the action and its label into atomic units significantly improves the performance of FSAR, improving 1-shot accuracy on HMDB51 by 4.1\% (79.9\% vs. 75.8\%) and 3.8\% (79.6\% vs. 75.8\%), respectively. 
In addition, the application of action decomposition in both the visual and textual domains improves performance by a further 5.0\%. 
This consistent improvements underscore the importance and effectiveness of atomic-level action decomposition for FSAR.
However, we observe that using visual or textual anatomy alone leads to performance degradation in the 5-shot setting.
We attribute this to the fact that in the 5-shot setting, sufficient samples in the support set allow the model to achieve effective classification. However, the addition of unimodal anatomy may introduce misalignment between different modalities, causing overfitting and degraded performance.
Furthermore, the additional video-text matching in our multimodal matching yields improvements of 6.0\%  and 1.1\%  under 1-shot and 5-shot settings, respectively, on HMDB51, confirming the effectiveness of multimodal matching.
Notably, the performance increase is particularly significant in the 1-shot scenario, highlighting the effectiveness of decomposing actions into atomic units and few-shot matching with different modality cues when visual information is limited. 
Moreover, the best performance is achieved by using both components together, empirically validating that multimodal fusion and multimodal matching complement each other.

\begin{table}[t]
\centering
\small
\caption{Comparison of different multimodal fusion schemes on the HMDB51 and SSv2-Small datasets with 5-way 1/5-shot settings. ``G'' and ``L'' denote global and local multimodal fusion between visual features and semantic cues, respectively.
}
\setlength{
    \tabcolsep}{
    0.7mm}{
\begin{tabular}
{c|cc|cc|cc}

\hline
			
\multirow{2}{*}{Text} &
\multirow{2}{*}{Scale} &
\multirow{2}{*}{Attention} &
\multicolumn{2}{c|}{{HMDB51}} &  
\multicolumn{2}{c}{{SSv2-Small}}    \\
 & & & \multicolumn{1}{l}{1-shot} & \multicolumn{1}{l|}{5-shot} & \multicolumn{1}{l}{1-shot} & \multicolumn{1}{l}{5-shot} \\ 
\shline

Action Label& G &Self & {79.3} & {85.9} & {58.0} & {66.1}  \\ 
Action Label& G &Cross & {81.7} & {85.9} & {58.5} & {67.2}  \\ 
\hline 
Atomic Description& G &Self & {85.2} & {87.6} & {58.4} & {65.4} \\ 
Atomic Description& L &Self & {85.5} & {88.8} & {57.8} & {66.5} \\ 
\rowcolor{Gray}
Atomic Description& L &Cross & {86.8} & {89.3} & {58.9} & {69.3} \\ 
\hline

\end{tabular}
}
\label{tab:network_fusion_ablation}
\vspace{-2mm}
\end{table}
\noindent\textbf{Analysis of Multi-modal Fusion.}
To better demonstrate the effectiveness of our Fine-grained Multimodal Fusion Module, we explore different multimodal fusion schemes to integrate visual features and semantic information.
As shown in the \cref{tab:network_fusion_ablation}, atomic action descriptions outperform action labels within the same fusion framework, as they provide more effective semantic information about the action. In addition, the cross-attention mechanism efficiently exploits semantic information by computing the similarity between textual and visual modality features, allowing it to better align and fuse the prior knowledge across different modalities. 
This enables the construction of more effective action representations in few-shot scenarios.
Overall, compared to the global-scale fusion of visual and textual features in CLIP-FSAR \cite{wang2023clip}, our method integrates textual and visual features at a fine-grained level, thus achieving more efficient action prototype construction and better few-shot action recognition results.

\begin{table}[t]
\centering
\small
\caption{Comparison results of different few-shot matching strategies on the HMBD51 and SSv2-Small datasets.
}
\setlength{
    \tabcolsep}{
    1.0mm}{
\begin{tabular}
{cc|cc|cc}

\hline

\multicolumn{2}{c|}{{Matching}} &  
\multicolumn{2}{c|}{{HMDB51}} &  
\multicolumn{2}{c}{{SSv2-Small}}    \\
 \multicolumn{1}{c}{video-video}& \multicolumn{1}{c|}{video-text} &  \multicolumn{1}{l}{1-shot} & \multicolumn{1}{l|}{5-shot} & \multicolumn{1}{l}{1-shot} & \multicolumn{1}{l}{5-shot} \\ 
\shline

OTAM \citep{2020-OTAM} & - & 80.9 & 88.1 & 55.2  & 66.3  \\ 
Bi-MHM \cite{2022-HyRSM} & - & 81.1 & 88.2 & 54.8 & 66.1 \\
AB-MHM & {-} & 80.9  & 88.2 & {56.5} & 69.2  \\
\hline 
- & Action label & \multicolumn{2}{c|}{{72.5}} & \multicolumn{2}{c}{{35.6}} \\
- & Atomic description & \multicolumn{2}{c|}{{81.9}} & \multicolumn{2}{c}{{36.1}} \\
\hline
OTAM \citep{2020-OTAM} & Atomic description & {86.7} & {89.3} & {56.5} & {66.3}  \\ 
Bi-MHM \cite{2022-HyRSM} & Atomic description & {87.6} & {89.6} & {55.2} & {66.4} \\
AB-MHM & Atomic description & {86.8} & {89.3} & {58.9} & {69.3} \\

\hline
\end{tabular}
}
\label{tab:matching_ablation}
\vspace{-2mm}
\end{table}
\noindent\textbf{Analysis of matching mechanism.}
In this section, we explore the role of our Multimodal Matching Module within the LGA framework. As shown in \cref{tab:matching_ablation}, we perform ablation experiments by changing matching strategies on the HMDB51 and SSv2-Small datasets. On HMDB51, the performance of different video-video matching strategies is generally comparable, but the additional atomic descriptions in our method significantly improve the video-text matching branch compared to the action label. Conversely, on SSv2-Small, video-text matching yields limited performance gains, while our proposed AB-MHM outperforms the Bi-MHM method branch by 3.7\% and 2.9\% under the 1-shot and 5-shot settings, respectively. 
We attribute this to the fact that HMDB51 contains richer scene information, enabling more accurate classification through enriched textual cues, while SSv2-Small has a stronger temporal focus with less scene content, making atomic-based video-video matching more effective.
Overall, these results demonstrate that our Multimodal Matching Module delivers robust few-shot matching across different datasets and settings.

\begin{table}[t]
\centering
\small
\caption{Comparison of different Temporal Segment methods on the HMDB51 and SSv2-Small datasets.
}
\setlength{
    \tabcolsep}{
    2.0mm}{
\begin{tabular}
{c|cc|cc}

\hline
			
\multirow{2}{*}{Segment Method} &
\multicolumn{2}{c|}{{HMDB51}} &  
\multicolumn{2}{c}{{SSv2-Small}}    \\
&  \multicolumn{1}{l}{1-shot} & \multicolumn{1}{l|}{5-shot} & \multicolumn{1}{l}{1-shot} & \multicolumn{1}{l}{5-shot} \\ 
\shline


TW-FINCH \citep{TW-FINCH} & {82.9} & {86.2} & {56.4} & {66.2}  \\ 
HARD & \textbf{{87.3}} & \textbf{{89.3}} & {57.0} & \textbf{{69.3}}  \\ 
CLUSTER w/o overlap & {86.6} & {88.3} & {58.3} & {68.0}  \\ 
\rowcolor{Gray}
CLUSTER & {86.8} & \textbf{{89.3}} & \textbf{{58.9}} & \textbf{{69.3}}  \\ 
\hline

\end{tabular}
}
\label{tab:temporal_segment_ablation}
\vspace{-4mm}
\end{table}
\noindent\textbf{Analysis of segment methods in Visual Anatomy.}
In this section, we analyze the effect of different temporal segmentation strategies in our framework. TW-FINCH \cite{TW-FINCH} is a temporally-weighted hierarchical clustering algorithm based on a 1-nearest neighbor graph. We use another baseline HARD that uniformly divides the video into three segments of equal length with the same overlap as ours. As shown in \cref{tab:temporal_segment_ablation}, we observe that our CLUSTER-Segment method consistently outperforms other competitors.
Specifically, TW-FINCH faces challenges in FSAR due to the lack of clear feature boundaries between atomic actions.
Furthermore, compared to HARD and CLUSTER (without overlap), our proposed CLUSTER—Segment method better captures atomic actions of varying temporal lengths. This enables the construction of reliable prototypes based on atomic actions, leading to improved performance.

\noindent\textbf{Analysis of Atomic Action number.}
We analyze the impact of varying the number of atomic actions on model performance. As shown in \cref{tab:action_num}, performance improves as the number of atomic actions increases, peaking at 3, but declines when increased to 4. 
This suggests a balance between granularity and coherence: while finer segmentation enhances action understanding, excessive splitting may introduces excessive temporal overlap, which dilutes the core motion signals and degrades model performance.
Additionally, a larger number of atomic actions amplifies the risk of LLM hallucinations, leading to ambiguous descriptions and misalignment between visual and textual features (further analysis in \cref{Effect of the atomic action number on the generated descriptions}). Therefore, we default the number of atomic actions to 3, which aligns well with the initiation, progression, and conclusion phases of an action.

\begin{table}[t]
\centering
\small
\caption{Analysis of different the num of
 atmoic action on the HMDB51 and SSv2-Small datasets.
}
\setlength{
    \tabcolsep}{
    4.0mm}{
\begin{tabular}
{c|cc|cc}

\hline
			
\multirow{2}{*}{Number} &
\multicolumn{2}{c|}{{HMDB51}} &  
\multicolumn{2}{c}{{SSv2-Small}}    \\
&  \multicolumn{1}{l}{1-shot} & \multicolumn{1}{l|}{5-shot} & \multicolumn{1}{l}{1-shot} & \multicolumn{1}{l}{5-shot} \\ 
\shline


1 & {54.3} & {65.4} & {86.7} & {88.7}  \\ 

2& 57.5 & 67.3 &87.6 &  89.3  \\ 

\rowcolor{Gray}
3 & {58.9} & {69.3} & {86.8} & {89.3}  \\ 

4 & 53.6 & 62.7 & 73.9 & 82.5 \\ 
\hline

\end{tabular}
}
\label{tab:action_num}
\vspace{-2mm}
\end{table}

\begin{figure}[t]
  \centering
  \includegraphics[width=\linewidth]{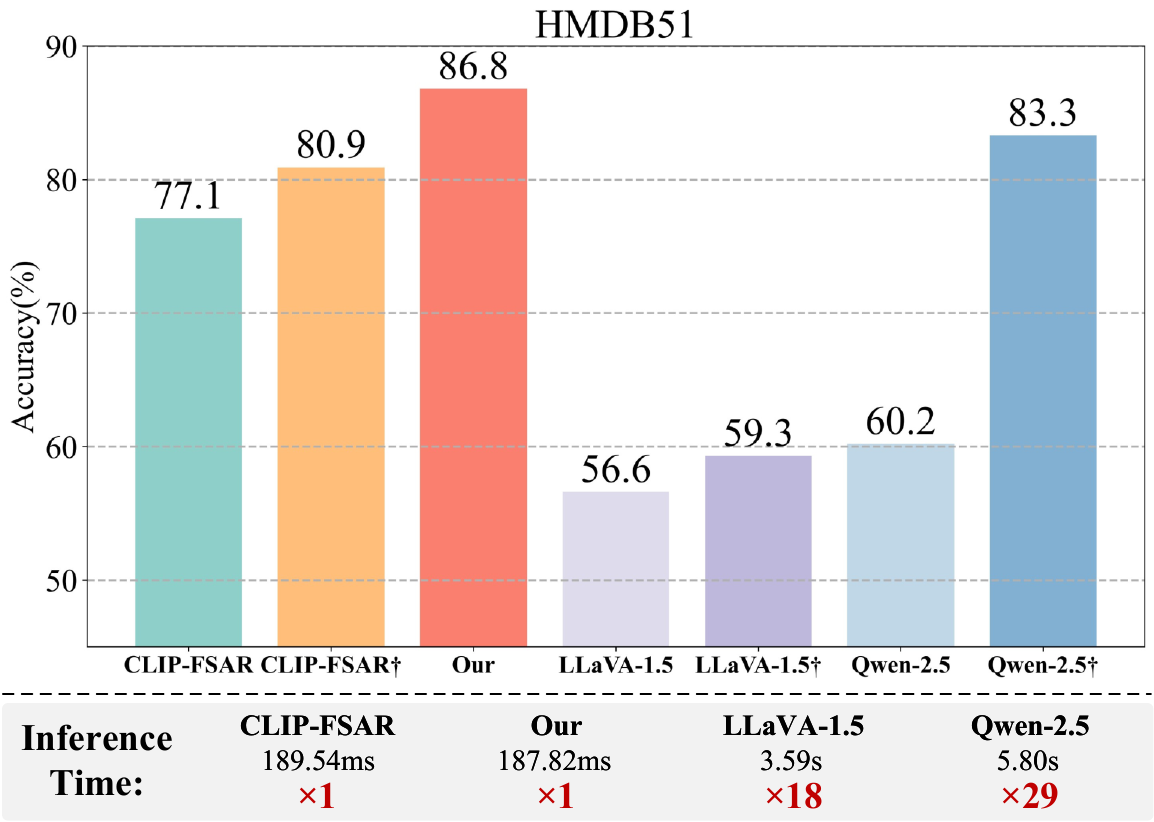}
   \caption{ Comparison experiments on the effect of LLM and VLM on the HMDB51 dataset.
    }
   \label{fig:VLM}
   \vspace{-4mm}
\end{figure}

\noindent\textbf{Analysis of pre-trained LLM and VLM}
In this section, we explore the impact of atomic action descriptions generated by LLM on existing FSAR methods. Specifically, we replace the action labels in CLIP-FSAR \cite{wang2023clip} with atomic action descriptions generated by the LLM model, resulting in CLIP-FSAR$\dagger$. As shown in \cref{fig:VLM}, the use of atomic action descriptions improves the performance of CLIP-FSAR, indicating that enriched action descriptions effectively improve the FSAR model's ability to represent actions in few-shot scenarios. 
In addition, we investigate the performance of pre-trained VLMs in the few-shot action recognition task. Specifically, we evaluate the LLaVA-1.5-7B and Qwen-2.5-7B models (details in supplementary). As shown in \cref{fig:VLM}, Qwen-2.5 slightly outperforms LLaVA-1.5 in zero-shot settings, but its performance remains significantly below that of mainstream FSAR methods. 
Furthermore, we explore the incorporation of label information into the prompt to obtain LLaVA-1.5$\dagger$ and Qwen-2.5$\dagger$, enabling further evaluation of VLM performance.
Although Qwen-2.5$\dagger$ outperforms most mainstream FSAR methods, the high computational requirements of VLM networks lead to significantly increase inference time and hinder its scalability across different N-way K-shot settings.
In contrast, the proposed LGA effectively leverages LLMs to explore action prior knowledge, thereby improving model performance while maintaining high inference efficiency.

\begin{figure}[t]
  \centering
  \includegraphics[width=\linewidth]{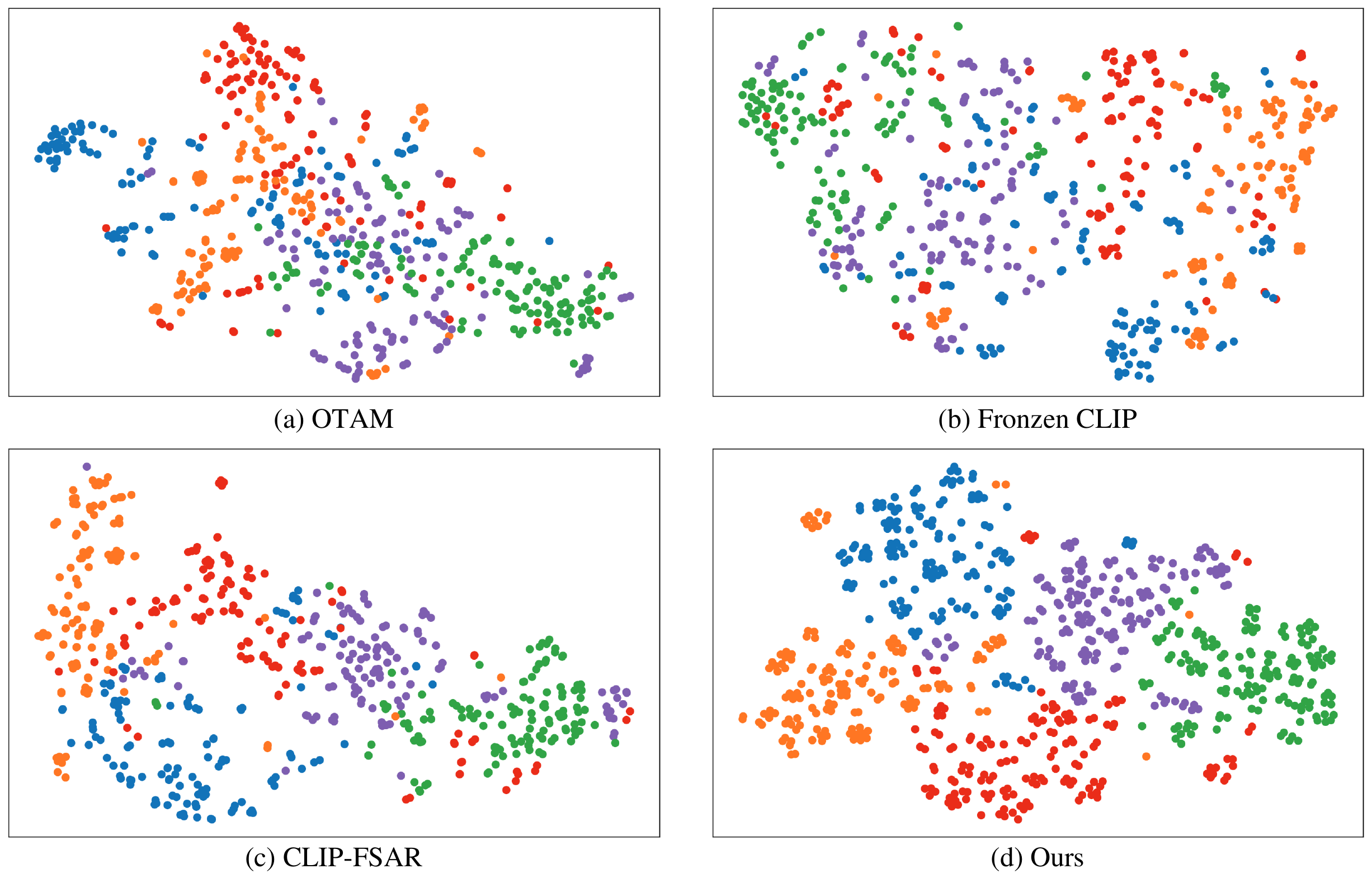}
   \caption{T-SNE distribution visualization of five action classes on the test set of SSv2-Small dataset. The different color represents video from different categories.}
   \label{fig:ablation_tsne}
   \vspace{-4mm}
\end{figure}

\subsection{Visualizations}
\textbf{Visualization of Prototype Similarity}.
To further analyze our method, we visualize the feature distribution of the baseline methods \cite{2020-OTAM, CLIP, wang2023clip} and our method in the test phase. 
As shown in \cref{fig:ablation_tsne}, we observe that our approach significantly improves the discriminative capabilities of action video feature representations compared to previous methods. 
Specifically, due to the limited training data and temporal dynamic information in the SSv2-Small dataset, OTAM \cite{2020-OTAM} and Frozen-CLIP \cite{CLIP} fail to produce accurate feature distributions for different action categories. 
While CLIP-FSAR \cite{wang2023clip} shows some improvement, demonstrating the effectiveness of incorporating multimodal information in few-shot action recognition, it still suffers from significant feature distribution overlap. 
In contrast, our method effectively captures discriminative features for different actions, resulting in a more accurate feature distribution and improved prediction accuracy.

\section{Conclusion}
We introduce Language-Guided Action Anatomy (LGA), a novel framework for few-shot action recognition that models actions at a fine-grained level by anatomizing both visual and textual modalities.
Unlike prior methods that rely on coarse action labels, LGA structures actions into atomic phases, capturing subject, motion, and object interactions during the initiation, progression, and conclusion of the action.
We integrate textual and visual features at the atomic level to generate more generalizable prototypes, and we present a Multimodal Matching Module for robust few-shot matching.
Extensive experiments on five widely used benchmarks demonstrate the effectiveness of our approach, with LGA achieving state-of-the-art performance.
\section{Acknowledge}
This work was partly funded by the Shanghai Municipal Science and Technology Major Project (2021SHZDZX0102), and STCSM (22DZ2229005).
{
    \small
    \bibliographystyle{ieeenat_fullname}
    \bibliography{main}
}

\clearpage
\setcounter{page}{1}
\maketitlesupplementary

\section{Prompt Design}
In our approach, to exploit the rich prior knowledge embedded in semantic space, we decompose each action label into an ordered sequence of atomic action descriptions with Visual Anatomy Module. Specifically, we leverage a large language model (GPT-4o) to transform action labels into atomic descriptions enriched with spatiotemporal context, as described in \cref{sec:action anatomy}. To ensure accurate parsing, we carefully design the prompt and extend it with additional input-output templates. The complete prompt as following:

\textbf{Prompt:} \textit{Deduce the scene description and three sub-action descriptions from an action label. The scene description should include possible scene elements, such as humans, objects, and background. The scene description must consist of visible elements, not abstract descriptions like atmosphere, mood or social setting. The sub-action descriptions must follow strict temporal order, focusing on the posture of the people involved, relevant elements in the scene, and potential interactive objects. Ignore object textures and dismiss any unlikely or invalid sub-actions, as well as unnecessary emotional descriptions. Keep the sub-action descriptions brief and clear and avoid the abstract descriptions such as enjoying the performance. Provide a concise answer for both the scene description and the three sub-action descriptions, following the example below:}

\textit{Example:
Input: jumping into pool.
Output: {``Action Label": ``Jumping into poo", ``sub-action description": [``A photo of a person stands at the edge of a pool, preparing to jump in.", ``A photo of a person leaps off the edge, mid-air over the pool.", ``A photo of a person enters the water, creating a splash as they dive in."]}
Your analysis should be thorough and accurate, considering all relevant aspects of the action to support your deductions effectively. Once I provide the action label, please deduce the scene description and three sub-action descriptions accordingly.}

\begin{figure*}[t]
  \centering
  \includegraphics[width=\linewidth]{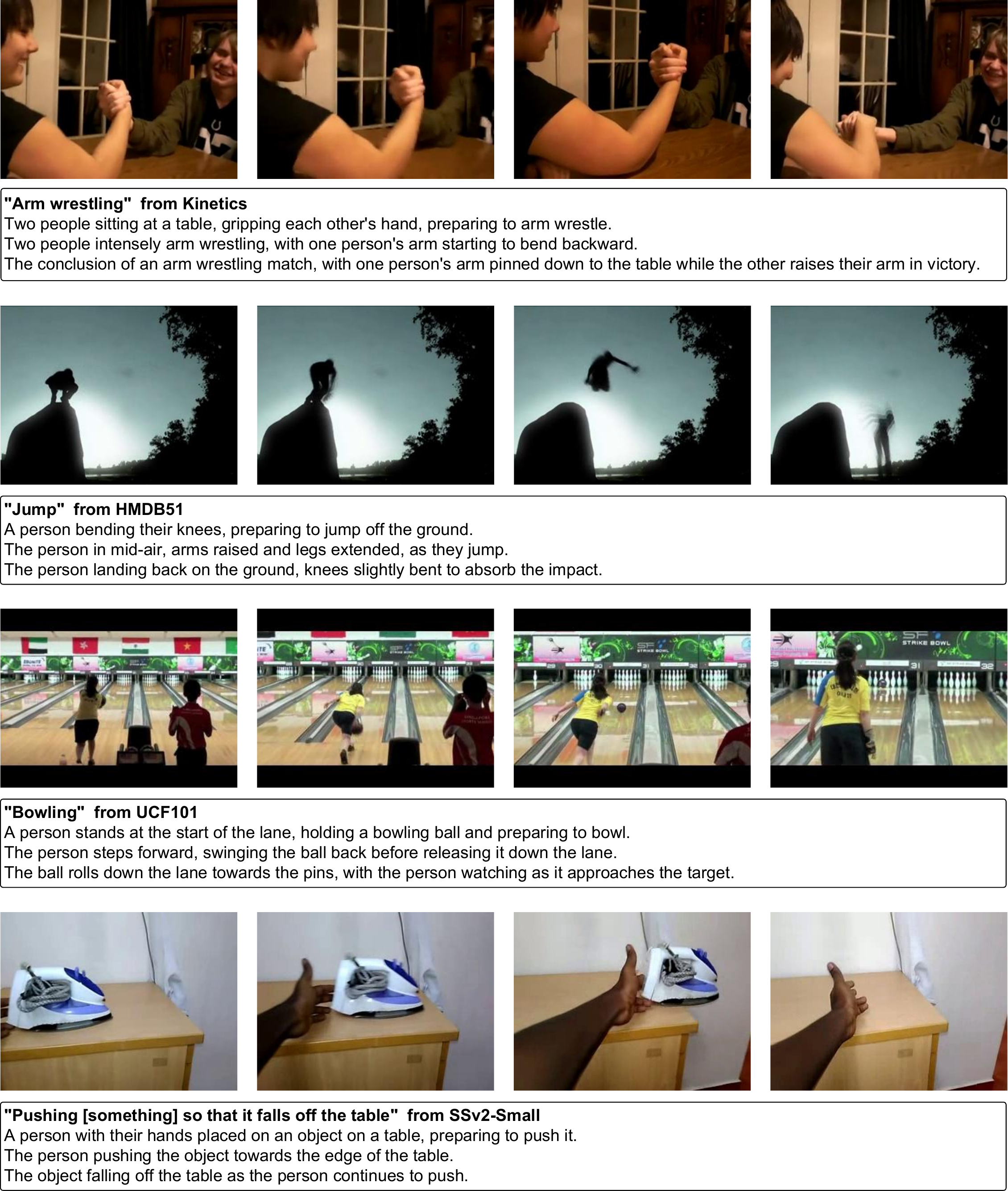}
   \caption{Atomic action descriptions generated from different datasets along with their corresponding action samples.}
   \label{fig:supp_prompt}
\end{figure*}

The output examples are shown in \cref{fig:supp_prompt}. The atomic action descriptions generated by the LLM effectively characterize the actions across different datasets.

\section{Implementation Details of Experimental}

\subsection{Network Parameters}
The hyperparameters of our methods in each dataset are shown in \cref{tab:hyperparameter}. In this table, `lr' means the learning tate, `st\_iter' indicates the number of iteration per step, `steps' refers to the number of steps to change the learning rate when using the multistep scheduler.
$\mathbf{M}$ and $\alpha$ means the attention mask weight in Fine-grained Multimdoal Fusion Module and the visual weight in Multimodal Matching Module, respectively.  
\begin{table}[t]
\centering
\small
\caption{The settings of hyperparameters in each dataset.}
\setlength{
    \tabcolsep}{
    0.7mm}{
\begin{tabular}
{c|c|c|c|c|c}

\hline
			
\multirow{1}{*}{Dataset} &
\multirow{1}{*}{lr} &
\multirow{1}{*}{st\_iter} &
\multirow{1}{*}{steps} &
\multirow{1}{*}{warm\_lr} &
\multirow{1}{*}{$\alpha$} 

\\

\shline

HMDB51 & 5e-6 & 700 & [0, 4, 6]   & 1e-6   & 0.0250 \\ 
Kinetics & 1e-5 & 250 & [0, 6, 9]   & 5e-6   &  0.0625\\ 
UCF101 & 2e-6 & 600 & [0, 4, 6]   & 1e-7 &    0.1125\\ 
SSv2-Small & 1e-5 & 2000 & [0, 4, 6]   & 1e-6 &  0.2 \\ 
SSv2-Full & 2e-6 & 600 & [0, 4, 6]   & 1e-7 &  0.2 \\ 

\shline
\end{tabular}
}
\label{tab:hyperparameter}
\vspace{-2mm}
\end{table}

\subsection{Effect of the atomic action number on the generated descriptions}
\label{Effect of the atomic action number on the generated descriptions}
In \cref{sec:ablation}, we have explored the impact of the number of atomic actions on the model. In this section, we further analyze how the number of atomic actions affects the generated atomic action descriptions.

\begin{figure*}[h]
  \centering
  \includegraphics[width=\linewidth]{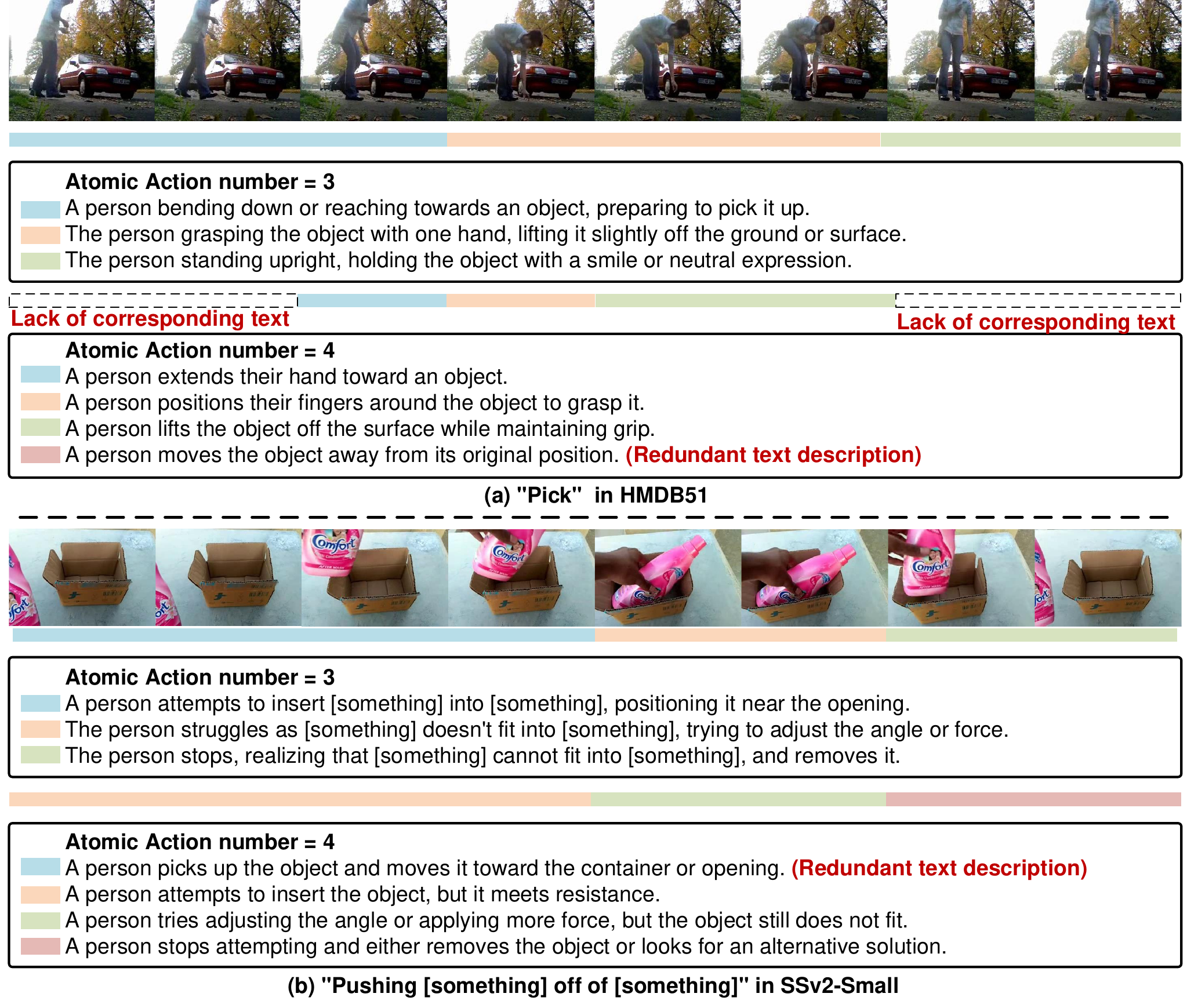}
   \caption{Atomic action descriptions generated with different numbers of atomic actions. (a) The action `pick' from HMDB51. (b) The action `Pushing [something] off of [something]' from SSv2-Small.}
   \label{fig:supp_action_num}
\end{figure*}

As shown in the \cref{fig:supp_action_num}, when the number of atomic actions is set to 3, the generated atomic action descriptions correspond well with the action samples. However, when the number is increased to 4, redundant atomic action descriptions are often generated. This redundancy can lead to misalignment between visual and textual features, which degrades model performance, as shown in the \cref{tab:action_num}.

\subsection{Prompt Design of VLM Experiments}
In \cref{sec:ablation}, we have explored the performance of pre-trained VLM models on FSAR tasks. Specifically, for LLaVA-1.5 and Qwen-2.5, we prompted the models with: \textit{``Below is a sequence of images showing an action. What action is being performed?"} to guide them in describing the query action. Finally, we classify the query action by measuring the distance between the semantic features of the description generated  by the VLM and  the labels of the support set.

Furthermore, for LLaVA-1.5$\dagger$ and Qwen-2.5$\dagger$, we incorporate support labels directly into the prompt to guide the VLM in classifying actions with support set labels. Specifically, we use the following prompt:
\textit{``Classify the action in the image sequence. Choose from: $\{$Action label1$\}$, $\{$Action label2$\}$, $\{$Action label3$\}$, $\{$Action label4$\}$, $\{$Action label5$\}$. Only output the action name from the list.}"
In this way, the VLM can classify actions within a limited set of labels, effectively mitigating the potential hallucinations associated with generating action descriptions.

\section{Additional Experiment Results}

\begin{figure*}[t]
  \centering
  \includegraphics[width=\linewidth]{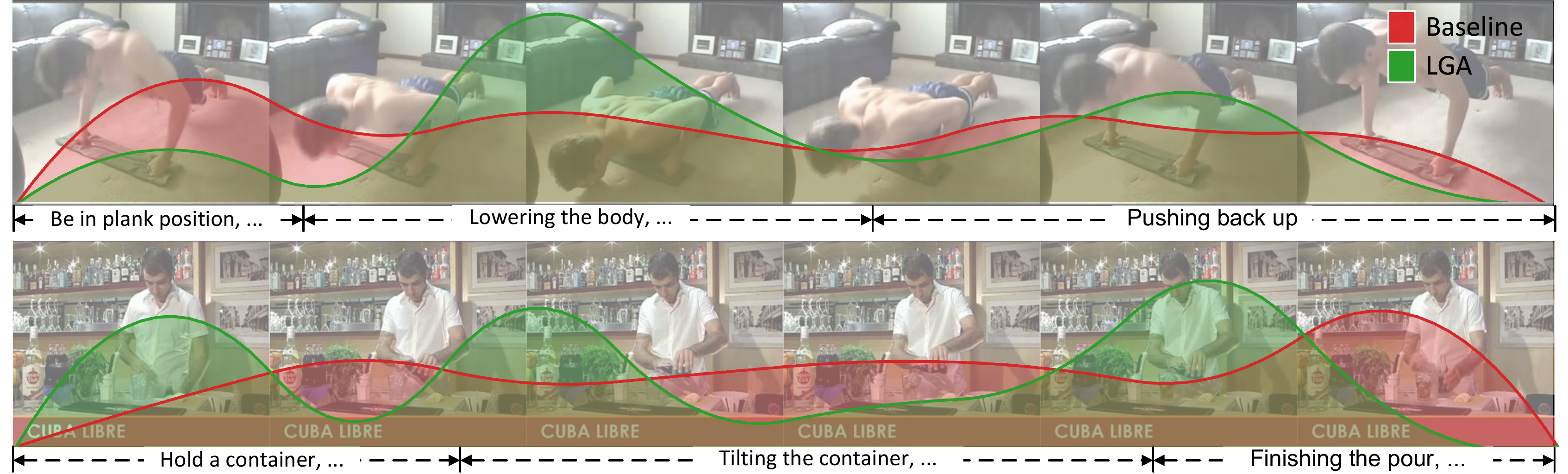}
   \caption{Temporal attention visualization of our LGA on HMDB51 \cite{kuehne2011hmdb51}.}
   \label{fig:supp_attn}
   \vspace{-4mm}
\end{figure*}

\subsection{Visualization of LGA}
To further evaluate LGA's ability to perceive the different temporal phases of actions, we visualize how each frame contributes to the overall distance between the video and the action prototype in HMDB51 \cite{kuehne2011hmdb51}.
As shown in \cref{fig:supp_attn}, compared to baseline methods, key frames from distinct temporal phases contribute more significantly to prototype matching.
For example, the frame depicting ``lowering the body'' in the ``Push up" sample contributes more significantly than baseline methods do.
This confirms that LGA selectively focuses on critical action phases rather than treating all frames uniformly.

\begin{figure}[t]
  \centering
  \includegraphics[width=\linewidth]{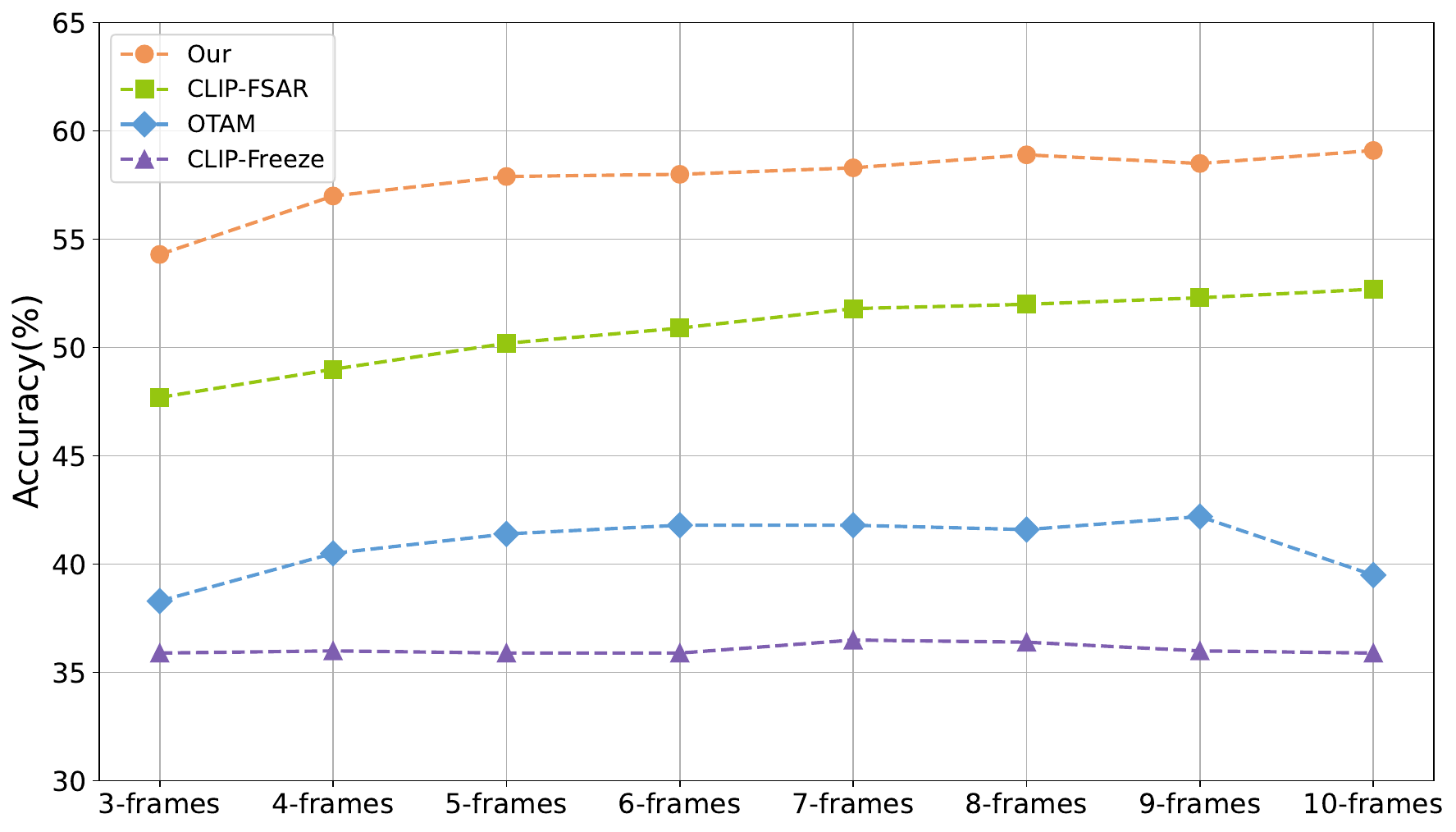}
   \caption{Performance comparison with different numbers of input video frames under 5-way 1-shot setting on SSv2-Small datasets.}
   \label{fig:supp_n_frame}
\end{figure}


\begin{table}[t]
\centering
\small
\caption{Comparison with recent state-of-the-art few-shot action recognition methods on the HMDB51 dataset with 5-way k-shot setting.
The best results are \textbf{bolded} in black, and the \underline{underline} represents the second best result.
%
}

\setlength{
    \tabcolsep}{
    2.0mm}{
\begin{tabular}
{l|c|ccc}
\hline
			
\hspace{0.0mm}  \multirow{2}{*}{Method} \hspace{2mm} 
& \multicolumn{1}{c|}{\multirow{2}{*}{Pre-training}} 
& \multicolumn{3}{c}{{HMDB51}}   \\
& \multicolumn{1}{c|}{} 
& \multicolumn{1}{l}{1-shot} & 3-shot  & 5-shot  \\
\shline

\shline
OTAM~\cite{2020-OTAM}  & CLIP-ViT-B
&72.5 &81.6 &83.9   \\

CLIP-Freeze~\citep{CLIP}   & CLIP-ViT-B
&58.2 &72.7 &77.0   \\ 

CapFSAR~\citep{CapFSAR} & BLIP-ViT-B
&65.2 & - & 78.6  \\

CLIP-FSAR~\cite{wang2023clip}  & CLIP-ViT-B
&77.1 &\underline{84.1} &87.7    \\ 

MA-FSAR ~\citep{xing2025mafsar} & CLIP-ViT-B
&{83.4}  & - &  {87.9}   \\  

Task-Adapter ~\citep{cao2024task-adpter} & CLIP-ViT-B
&{83.6}  & - &  {88.8}   \\  

EMP-Net ~\citep{wu2025empnet} & CLIP-ViT-B
&76.8  & - &  85.8   \\  

Kumar\textit{et al.}~\citep{kumar2024trajectory} & DINOv2
&60.0  & 71.8 &  77.0    \\ 

CLIP-CPM$^2$C~\citep{guo2025clipcpmc}& CLIP-ViT-B
&75.9  & - & 88.0   \\

TSAM ~\citep{li2025TSAM} & CLIP-ViT-B
&\underline{84.5}  & - & \underline{88.9}   \\

\rowcolor{Gray}
Our~   & CLIP-ViT-B
& \textbf{86.8}   & \textbf{89.2}  & \textbf{89.3}  \\

%
\hline
\end{tabular}
}

\label{tab:compare_supp_k_shot}
\vspace{-2mm}
\end{table}
\subsection{Analysis of the number of support set samples}
In this section, we further evaluate the performance of our method under different shot settings. As shown in \cref{tab:compare_supp_k_shot}, our method achieves greater improvements over the baseline method (CLIP-FSAR \citep{wang2023clip})  in lower-shot scenarios; specifically, it outperforms the baseline by 9.7\% in the 1-shot setting compared to 1.6\% in the 5-shot setting. We attribute this to the fact that enriched semantic cues are particularly effective when visual information is limited.

\subsection{Analysis of the number of input video frames}
To ensure a fair comparison with previous methods \cite{2020-OTAM, 2022-HyRSM, 2023-molo, huang2024matching}, we uniformly sample 8 frames from each video to construct its visual representation in our experiments. In addition, to comprehensively analyze the influence of the number of frames on the model performance, we perform an ablation study on the HMDB51 and SSv2-Small datasets by varying the number of input frames. As shown in \cref{fig:supp_n_frame}, the performance starts to increase and gradually saturates as the number of input frames increases.  Remarkably, our method outperforms previous methods across different frame number settings, demonstrating its effectiveness and robustness. 

\begin{figure}[t]
  \centering
  \includegraphics[width=\linewidth]{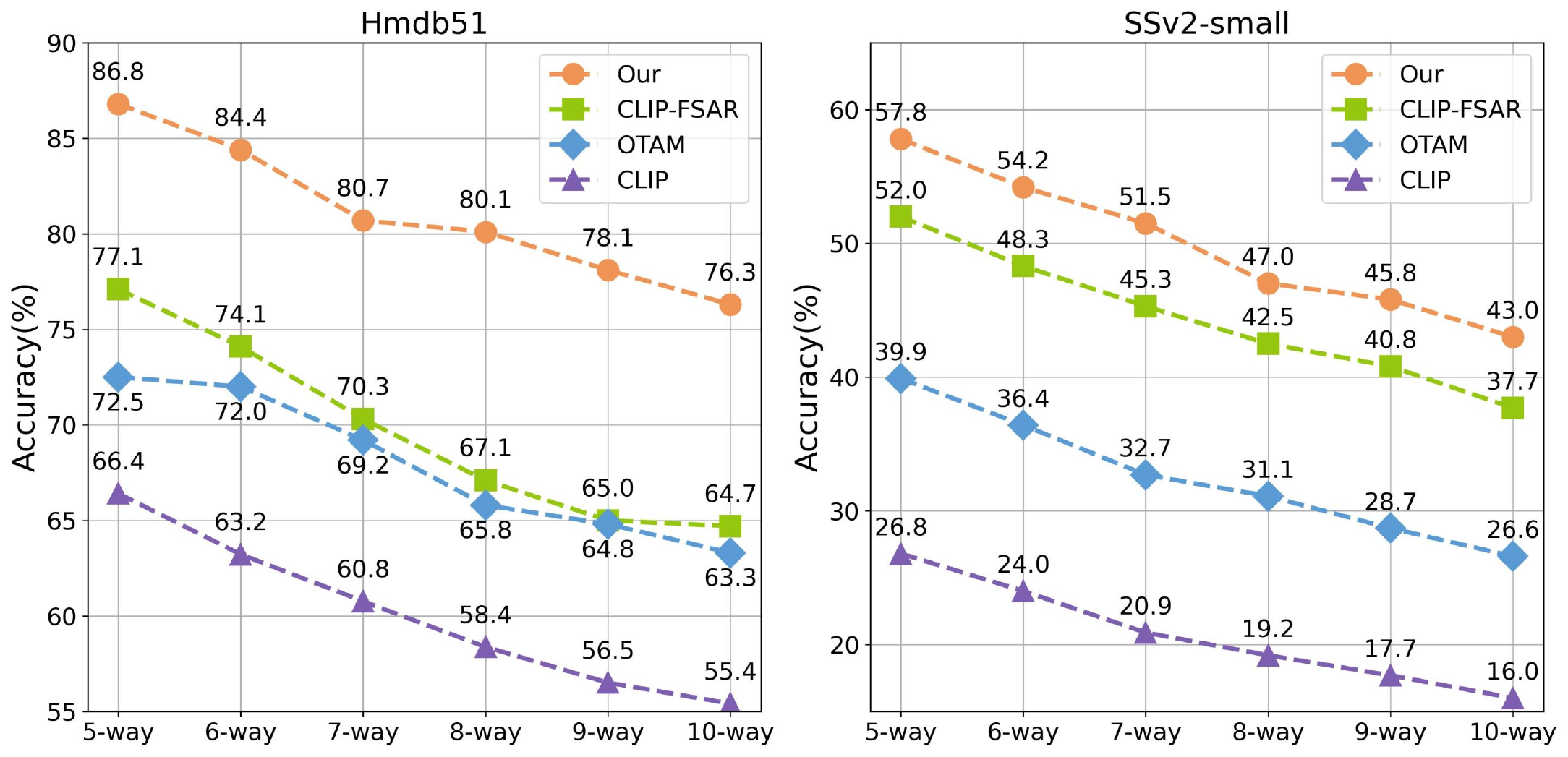}
   \caption{$N$-way 1-shot results of our method and other baseline methods with $N$ varying from 5 to 10.}
   \label{fig:ablation_n_way}
   \vspace{-4mm}
\end{figure}

\subsection{Analysis of N-way classification}
We also investigate the effect of varying $\mathnormal{N}$ on the few-shot performance. We further perform ablation experiments to evaluate the $\mathnormal{N}$-way 1-shot accuracy, where $\mathnormal{N}$ ranges from 5 to 10. As shown in \cref{fig:ablation_n_way}, we can observe that as $\mathnormal{N}$ increases, the classification becomes more challenging, resulting in a noticeable drop in performance. Specifically, on the SSv2-Small dataset, our method experiences a 14.8\% drop in accuracy when moving from the 5-way 1-shot setting to the 10-way 1-shot setting. Despite this performance drop, our approach consistently outperforms all baseline methods in all evaluated settings, highlighting the robustness and superiority of our method.


\end{document}